\documentclass[10pt,twocolumn,letterpaper]{article}

\usepackage{cvpr}              

\usepackage[dvipsnames]{xcolor}
\usepackage{times}
\usepackage{epsfig}
\usepackage{color, colortbl}
\usepackage{mathtools}
\usepackage{multirow}
\usepackage{makecell}
\usepackage[accsupp]{axessibility}
\usepackage{dsfont}
\usepackage{nicematrix}

\definecolor{highlight}{RGB}{236, 236, 236}
\usepackage{algorithm2e}

\newcommand{\best}[1]{{\textbf{#1}}}

\newcommand{\ourname}[1]{HandBooster}

\definecolor{cvprblue}{rgb}{0.21,0.49,0.74}
\usepackage[pagebackref,breaklinks,colorlinks,citecolor=cvprblue]{hyperref}

\def\paperID{6065} 
\def\confName{CVPR}
\def\confYear{2024}

\title{
\ourname{}: Boosting 3D Hand-Mesh Reconstruction by \\Conditional Synthesis and Sampling of Hand-Object Interactions
}

\author{Hao Xu$^{1}$ \hspace{0.5cm} 
Haipeng Li$^{2}$ \hspace{0.5cm}
Yinqiao Wang$^{1}$ \hspace{0.5cm}
Shuaicheng Liu$^{2}$ \hspace{0.5cm}
Chi-Wing Fu$^{1}$ \\
\textsuperscript{1}The Chinese University of Hong Kong\\
\textsuperscript{2}University of Electronic Science and Technology of China\\
{\tt\small \{xuhao,yqwang,cwfu\}@cse.cuhk.edu.hk, \{lihaipeng@std.,liushuaicheng@\}uestc.edu.cn}
}

\begin{document}
\maketitle

\footnotetext[1]{Department of Computer Science and Engineering; Institute of Medical Intelligence and XR (IMIXR).}

\begin{abstract}
Reconstructing 3D hand mesh robustly from a single image is very challenging, due to the lack of diversity in existing real-world datasets.
While data synthesis helps relieve the issue, the syn-to-real gap still hinders its usage.
In this work, we present \ourname{}, a new approach to uplift the data diversity and boost the 3D hand-mesh reconstruction performance by training a conditional generative space on hand-object interactions and purposely sampling the space to synthesize effective data samples.
First, we construct versatile content-aware conditions to guide a diffusion model to produce realistic images with diverse hand appearances, poses, views, and backgrounds; favorably, accurate 3D annotations are obtained for free. 
Then, we design a novel condition creator based on our similarity-aware distribution sampling strategies to deliberately find novel and realistic interaction poses that are distinctive from the training set.
Equipped with our method, several baselines can be significantly improved beyond the SOTA on the HO3D and DexYCB benchmarks.
Our code will be released on {\small \url{https://github.com/hxwork/HandBooster_Pytorch}}.

\end{abstract}
    
\vspace{-4mm}
\section{Introduction}
\label{sec1:intro}

The task of reconstructing 3D hand mesh from a single image facilitates a wide range of applications,~\eg, in AR/VR and human-computer interactions. Recently-proposed data-driven methods show promising results in hand-object interaction scenarios.
Yet, their performance is largely limited by the training data, since existing datasets typically lack diversity in hand appearances/poses, views,~\etc

Existing real-world hand-object datasets are collected in laboratory or in-the-wild scenes.
For laboratory-captured datasets such as DexYCB~\cite{chao2021dexycb} and HO3D~\cite{hampali2020honnotate},
they offer a large quantity of hand-object interaction samples with relatively accurate 3D annotations.
However, the variations in the samples are still limited, since the data collection process is typically very tedious with the expensive MoCap system.
Conversely, for in-the-wild datasets,~\eg, YouTube-Hands~\cite{kulon2020weakly}, the data has richer variations, but they provide only pseudo labels, 
without accuracy guarantee.

\begin{figure}[t]
    \centering
    \includegraphics[width=\linewidth]{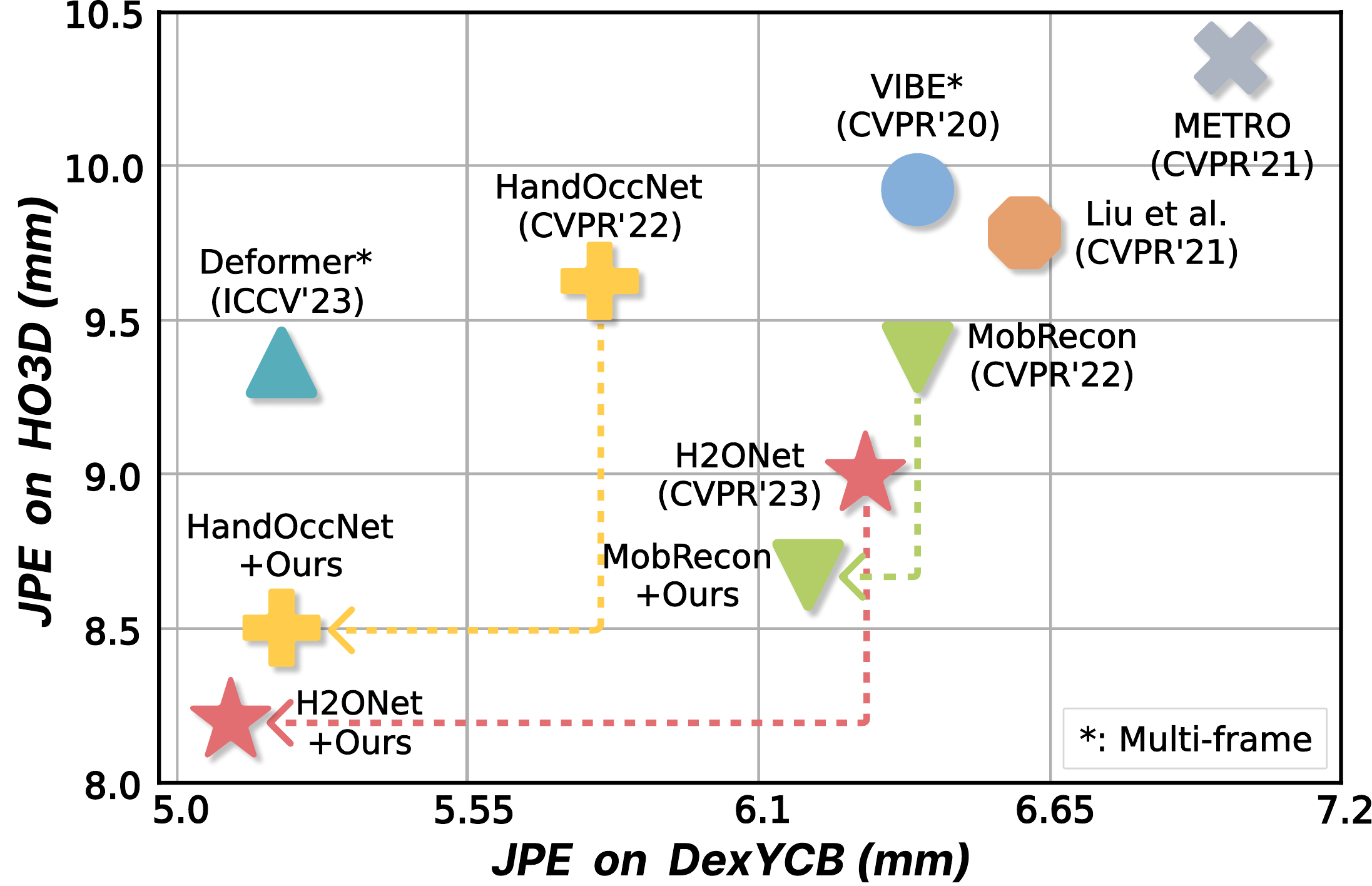}
    \vspace*{-6mm}
    \caption{\ourname{} significantly improves several baselines, making them SOTA again on both HO3D and DexYCB.
    }
    \label{fig:teaser}
    \vspace{-5mm}
\end{figure}

Synthetic data, both rendering- and generative-based, is a workaround to avoid tedious data collection while ensuring annotation precision.
ObMan~\cite{hasson2019learning}, YCB-Afford~\cite{corona2020ganhand}, and ArtiBoost~\cite{yang2022artiboost} use Blender~\cite{blender}, Maya~\cite{maya}, or PyRender~\cite{pyrender} to produce hand renderings.
However, the results are unrealistic due to simulated hand appearance and inconsistency in foreground/background lighting.
Also, they need extensive extra data such as HDR environment maps for lighting and background augmentation. 
On the other hand, generative methods,~\eg, HOGAN~\cite{hu2022hand}, can alleviate these issues.
Yet, they consider only novel view synthesis but not other aspects of data diversity.
Also, there is no evidence that the hand-mesh reconstruction performance can be improved consistently on existing methods.
To sum up, the key challenge is how to generate {\em realistic and diverse hand-object interaction images with reliable annotations\/}.

In this paper, we present \ourname{} to uplift the data diversity and boost 3D hand-mesh reconstruction {\em by training a conditional generative space on hand-object interactions\/} and {\em purposely sampling the space to produce effective data samples\/}.
In short, we first construct versatile content-aware conditions to guide a diffusion model to produce realistic image samples.
By this means, we can controllably produce images of diverse hand appearances, poses, views, and backgrounds, where precise 3D annotations are available for free. 
Further, we design a novel condition creator based on our similarity-aware distribution sampling strategies to find novel interaction poses distinct from the existing ones, thus maximizing the training data quality.

First, directly mapping an input 3D mesh to 2D RGB values is challenging.
So, we decompose this process into two steps:
(i) project the 3D mesh into a more interpretable 2D image form while preserving its geometric information; and 
(ii) utilize the 2D results as conditions on the diffusion model to enable controllable generation of realistic images.
Among the candidates for the 2D conditions, inspired by~\cite{hu2022hand}, we empirically choose the comprehensible and informative normal map and object texture. 
Further, observing that small changes in 3D hand orientation can hardly be seen in 2D images, we embed 3D hand orientation as an additional condition to ensure orientation-aware generation.

Second, to enhance the reconstruction performance and generalizability, solely utilizing existing hand-object grasping poses is inadequate.
The synthesized hand-object interaction samples should be: (i) realistic, ensuring natural grasping poses; (ii) diverse, encompassing various grasping poses; and (iii) novel, encouraging unseen poses. 
To achieve (i), we employ an optimization-based method~\cite{wang2023dexgraspnet} to simulate grasping poses, followed by our validation strategy to find natural poses.
To promote (ii) and (iii), we further propose similarity-aware sampling strategies, including an intra-distribution farthest pose sampling strategy to avoid repeated or similar poses within both the real and synthetic data and a cross-distribution sampling strategy to encourage the likelihood of sampling novel-generated poses.
Last, we train several baselines using our generated samples together with real-world data, showing that their performance can be significantly improved over the SOTA; see Fig.~\ref{fig:teaser}.

Our main contributions are summarized as follows,
\begin{itemize}
\item
We propose \textbf{\ourname{}}, a new generative framework that utilizes content-aware conditions to synthesize realistic hand-object images with diverse hand appearances, poses, views, and backgrounds, as well as accurate annotations, to boost the reconstruction performance.
\item
We design the novel condition creator to effectively produce realistic and diverse novel views and grasping poses, by designing the intra-distribution farthest pose sampling strategy and the cross-distribution sampling strategy.
\item
Extensive evaluations on two widely-used datasets show that \ourname{} achieves consistent performance gain on
several methods, setting new SOTA performance.
\end{itemize}

\section{Related Works}
\label{sec2:rw}
\paragraph{3D Hand-Mesh Reconstruction.}
Extensive research has been conducted on 3D hand-mesh reconstruction from RGB images.
Most of them tackle the problem by regressing MANO coefficients~\cite{zhang2019end, zhou2020monocular, yang2020bihand, hasson2019learning, zhang2021hand, boukhayma20193d, baek2019pushing, zimmermann2019freihand, chen2021model, zhao2021travelnet, cao2021reconstructing, baek2020weakly, jiang2021hand, liu2021semi, zhang2021interacting, tse2022collaborative}.
Others mainly regress voxels~\cite{iqbal2018hand, moon2020i2l, moon2020interhand2, yang2021semihand}, implicit functions~\cite{mescheder2019occupancy}, and vertices~\cite{ge20193d, kulon2020weakly, chen2021camera, lin2021end, lin2021mesh, chen2022mobrecon}. 
Despite careful network designs, reconstructing accurate 3D hand meshes from monocular images remains challenging, particularly under severe occlusions.
Hence, some recent works~\cite{yang2020seqhand, chen2021temporal, xu2023h2onet, wen2023hierarchical, fu2023deformer} attempt to leverage multi-frame information.
A very recent work~\cite{ye2023diffusion} uses prior knowledge in a diffusion model to render object geometries to improve everyday object reconstruction.
Our approach is orthogonal to these methods and achieves top performance when partnered with several recent baselines~\cite{park2022handoccnet, chen2022mobrecon, xu2023h2onet}.

\begin{figure*}[t]
    \centering
    \includegraphics[width=\linewidth]{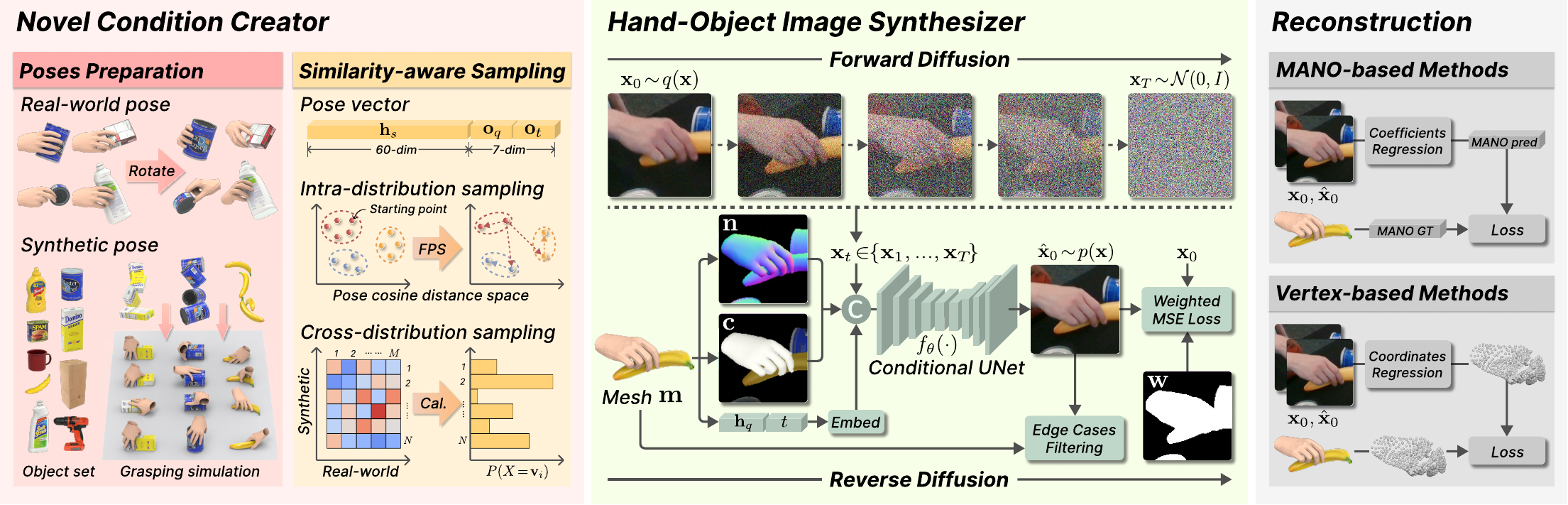}
    \vspace*{-6mm}
    \caption{Our \ourname{} framework.
    (i) The Novel Condition Creator prepares and samples diverse and novel grasping poses against real-world and synthetic distributions to create conditions.
    (ii) The Hand-Object Image Synthesizer follows the conditions to generate image samples.
    (iii) The synthesized samples can then be employed to effectively train different types of reconstruction models.}
    \label{fig:pipeline}
    \vspace{-4mm}
\end{figure*}

\vspace{-3mm}
\paragraph{Hand-Object Interaction Image Synthesis.}
So far, few works have explored image synthesis of hand-object interactions.
These works are either rendering- or generative-based.
The former employs rendering tools such as Blender~\cite{blender} and Maya~\cite{maya} to produce synthetic data.
To generate grasping poses, some~\cite{corona2020ganhand,yang2022artiboost,jian2023affordpose} design specific algorithms, while others~\cite{hasson2019learning} use off-the-shell tools such as GraspIt~\cite{miller2004graspit} and recent works~\cite{wang2023dexgraspnet,xu2023unidexgrasp,wan2023unidexgrasp++}. 
However, when applied to 3D hand-mesh reconstruction, these synthetic data inevitably introduce noticeable domain gaps compared to real-world data, leading to inferior performance. 

Generative-based methods can produce more realistic images. HOGAN~\cite{hu2022hand} synthesizes novel views using the target posture as guidance; yet, it cannot effectively generate images of novel grasping poses and lacks other aspects of diversity. 
Another work~\cite{ye2023affordance} generates hand graspings on RGB images that contain objects by using a diffusion model conditioned on a generated hand orientation mask. 
However, its diversities are still restricted by the input images and the lack of annotations severely limits its usage in downstream applications.
In this work, we are able to synthesize realistic and diverse hand-object interactions that encompass various appearances, grasping poses, object types, and camera views, significantly enhancing the hand-mesh reconstruction performance, as shown in Fig.~\ref{fig:teaser}.

\vspace{-3mm}
\paragraph{Conditional Diffusion Model.}
Diffusion model is a type of generative model that adopts the stochastic diffusion process in thermodynamics~\cite{sohl2015deep}.
Besides, it can also be formulated as a score-based generative model~\cite{song2019generative, song2020score}. 
Recently, DDPM~\cite{ho2020denoising} models complex data distributions through discrete steps. 
In this work, we focus mainly on conditioned generation, including classifier-guided~\cite{dhariwal2021diffusion,liu2023more} and classifier-free~\cite{ho2022classifier} methods.
Though LDM~\cite{rombach2022high}, ControlNet~\cite{zhang2023adding}, or other related methods~\cite{wang2022zero, hu2022lora, song2020denoising, karras2022elucidating, lu2022dpm, bao2022analytic, salimans2022progressive, hoogeboom2023simple, li2024dmhomo, jiang2023low} appear as potential tools for our work, they are primarily designed for producing high-resolution images or exploiting data priors in pre-trained models, lacking task-specific designs to meet the needs of 3D hand-mesh reconstruction,~\eg, plausible hand-object interactions and hand orientation awareness.
In this work, we propose various content-aware conditions to guide the DDPM, demonstrating controllability and fine-grained realistic results.

\section{Method}
\label{sec3:method}

\subsection{Overview}
Fig.~\ref{fig:pipeline} shows the overall \ourname{} framework.
First, we formulate the hand-object image synthesizer, which utilizes content-aware conditions to generate compatible hand-object images (Sec.~\ref{sec3.2}).
Next, to maximize its efficacy, we design the novel condition creator, aiming to prepare and sample realistic, diverse, and novel grasping poses as conditions (Sec.~\ref{sec3.3}). 
Combining their strengths, we can effectively generate realistic hand-object images with annotations to train various reconstruction models (Sec.~\ref{sec3.4}.)

\subsection{Hand-Object Image Synthesizer}
\label{sec3.2}
We first describe the process of generating realistic images from a given 3D hand mesh $\mathbf{m}$ using our hand-object image synthesizer.
To tackle the challenge of constructing a mapping between the 3D coordinate space and the 2D image space, we decompose the process into two steps, as inspired by~\cite{rombach2022high, ho2022classifier}.
First, we perform a 3D-to-2D projection while retaining sufficient information to construct content-aware conditions.
Then, we adopt them to guide a conditional diffusion model to generate realistic hand-object images.

\begin{figure}[t]
    \centering
    \includegraphics[width=\linewidth]{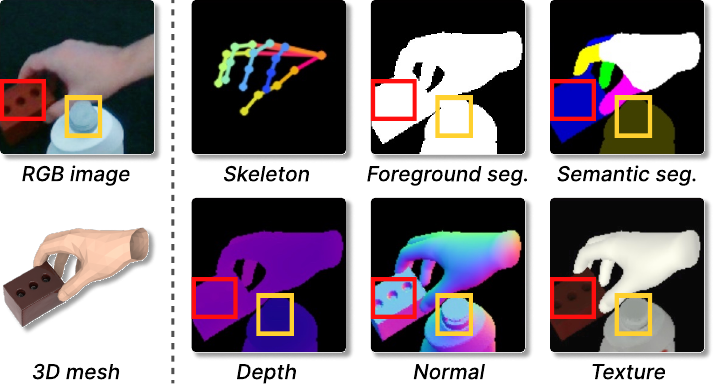}
    \vspace*{-6mm}
    \caption{Condition candidates. Normal and texture maps are more informative than others; see the red/yellow bounding boxes.}
    \label{fig:conditions}
    \vspace{-4mm}
\end{figure}

\vspace{-3mm}
\paragraph{Content-aware Conditions.}
To minimize the information lost during the 3D-to-2D projection process and reduce the learning difficulty, we aim to choose informative and interpretable conditions for synthesizing hand-object images.
We have considered several candidates in 2D format, including skeleton, segmentation, texture, depth, and normal maps, which can be divided into two categories: depth and normal maps contain more topology information, while others are more semantic. 
To strike a balance, a sound solution is to choose one from each category. 
As Fig.~\ref{fig:conditions} shows, the texture map contains both shape and color knowledge, so it is selected due to its high informativeness. 
Also, the normal map is selected for its ease of interpretation, as different parts of the hand and object are more distinguishable compared to the depth map. 
However, relying solely on 2D images cannot capture small changes in the 3D hand orientation, leading to large regression errors in the camera space during the reconstruction. 
Hence, we design another condition to encourage orientation-aware generation, where the hand orientation is represented using quaternion, embedded into latent space, and incorporated into several stages of the diffusion model. This inclusion allows for better handling of changes in hand orientation and gives hints to generate arms (not included in conditions) in the image synthesis.

\vspace{-3mm}
\paragraph{Controllable Realistic Image Synthesis.}
We first utilize the classifier-free diffusion model~\cite{ho2020denoising, ho2022classifier} for a controllable generation of hand-object images.
The forward diffusion is a Markov chain of diffusion steps, in which Gaussian noise is added progressively to a real-world data sample, $\mathbf{x}_0\!\!\sim\!\! q(\mathbf{x})$, producing a noisy transition sequence $\mathbf{x}_1,\mathbf{x}_2,...,\mathbf{x}_T$.
To facilitate the training, we use the parameterization method following~\cite{kingma2014adam} to sample $\mathbf{x}_t$ at arbitrary timestamp.
Then, we adopt a UNet-like classifier-free denoising model $f_\theta(\cdot)$ to learn the reverse diffusion, taking our content-aware conditions to control the generation of the hand-object images from an isotropic Gaussian noise.
After the training, we can obtain a controllable model for synthesizing high-quality images with fine-grained details, following our content-aware conditions.

During training, $\mathbf{x}_0\!\in\!\mathbb{R}^{128\!\times128\!\times\!3}$ are the same image used for 3D hand-mesh reconstruction. 
To improve the quality of the generated data and employ them in our task, we incorporate our content-aware conditions $\mathbf{y}$ with $\mathbf{x}_t$ and $t$ to generate samples $\hat{\mathbf{x}}_0$ using the denoising model $f_{\theta}(\cdot)$, 
\begin{equation}
\small
    \hat{\mathbf{x}}_0=f_\theta(\operatorname{Concat}(\mathbf{n},\mathbf{c}),\mathbf{h}_{q},\operatorname{PE}(t)),
    \vspace{-2mm}
\end{equation}
where $\operatorname{Concat}(\cdot)$ denotes the channel-wise concatenate operation;
$\mathbf{n}$ and $\mathbf{c}$ indicate the normal and texture maps, respectively; and 
$\mathbf{h}_{q}$ is the hand orientation, represented using quaternion to ensure continuity; and the timestamp $t$ is encoded through positional embedding (PE)~\cite{vaswani2017attention}. 
Following~\cite{karras2022elucidating}, we let the denoising model $f_\theta(\cdot)$ predict RGB values and calculate L2 loss with $\mathbf{x}_0$. Formally, we have
\begin{equation}
\small
    \mathcal{L}_{DM}=\mathbf{w}\frac{\Bar{\alpha}_t}{1-\Bar{\alpha}_t} ||\hat{\mathbf{x}}_0-\mathbf{x}_0||_2^2, 
    \vspace{-2mm}
\end{equation}
where $\mathbf{w}$ is a pixel-wise weighting map to enhance the foreground, since the generation quality of the background is less important. We use the hand-object segmentation map as the mask and set the weight to 0.1 for the background.

In the reverse diffusion, synthetic samples $\hat{\mathbf{x}}_0$ are generated from randomly-sampled Gaussian noise $\mathbf{x}_t\!\in\!\mathcal{N}(\mathbf{0},\mathbf{I})$ with the content-aware conditions $\mathbf{n}$, $\mathbf{c}$, and $\mathbf{h_{q}}$. After certain timestamps, the generated images $\hat{\mathbf{x}}_0$ would become realistic and the input 3D meshes $\mathbf{m}$ are exactly the corresponding ground truths, forming a training pair $(\hat{\mathbf{x}}_0, \mathbf{m})$.

\vspace{-3mm}
\paragraph{Edge Cases Filtering.} 
Though we can effectively generate realistic images, it is unlikely to fully avoid undesired artifacts, so filtering is necessary.
To mitigate the negative effects of training the reconstruction models with such image samples, we propose a selection process to filter out edge cases.
Specifically, we adopt a 3D hand-mesh reconstruction model pre-trained on the same training set to evaluate the generated images $\hat{\mathbf{x}}_0$. 
Here, we calculate the 3D joint and vertex errors between the predicted results and the corresponding ground truths $\mathbf{m}$. 
Considering that the model should yield reasonable outcomes on eligible synthetic images that align with the distribution of real-world data,
we thus exclude edge-case samples that exhibit significant errors over predetermined thresholds in the selection.

\subsection{Novel Condition Creator}
\label{sec3.3}
To effectively generalize the reconstruction models to handle unseen scenarios while avoiding potential over-fitting, it is essential to incorporate diverse hand-object interaction poses. 
However, simply augmenting existing conditions can only produce novel views.
Doing so is insufficient to boost the reconstruction performance. 
Hence, we design the novel condition creator to construct novel and diverse conditions by finding unseen poses and combining them with the existing ones via our distribution sampling strategies.

\vspace{-3mm}
\paragraph{Pose Preparation.} 
We first try to fully utilize the grasping poses in the current training set by augmenting their hand orientations. Since not all frames involve object grasping, the data is split into ``grasping'' and ``non-grasping'' parts. Only ``grasping'' poses are augmented to synthesize novel views, while others remain unchanged to preserve realistic motions. Yet, existing datasets~\cite{chao2021dexycb, hampali2020honnotate} do not provide labels for grasping status, so we derive the grasping status
from the object pose automatically. 
We calculate the isotropic Relative Rotation Error (RRE) and Relative Translation Error (RTE) between the initial and current poses:
\begin{equation}
\small
    \mathrm{RRE}\!=\!\arccos(\frac{\mathrm{trace}({\mathbf{o}_{r}^t}^T \mathbf{o}_{r}^{0} \!-\! 1)}{2}),\; \mathrm{RTE}\!=\!||{\mathbf{o}_{t}^{t}-\mathbf{o}_{t}^{0}}||_2,
\end{equation}
where $\mathbf{o}_{r}$ and $\mathbf{o}_{t}$ denote the object rotation matrix and translation vector, respectively; and 
the superscripts $0$ and $t$ denote the first and $t$-th frames, respectively, in the same sequence. If the difference exceeds a pre-defined threshold (empirically setting $5^{\circ}$ on RRE and $10mm$ on RTE), the frame is labeled as ``grasping''.
To produce novel views, the orientations of the ``grasping'' poses are perturbed in a certain range and then rendered as conditions.

Next, we focus on generating novel hand-object grasping poses. 
Specifically, we use the same YCB~\cite{calli2015ycb} objects as in DexYCB~\cite{chao2021dexycb} and obtain initial poses by simulating the process of falling from a random height to a plane. 
To generate a grasping pose, we employ the recent work~\cite{wang2023dexgraspnet} for its fast convergence and high success rate. Please refer to~\cite{xu2023unidexgrasp} for the details. 
We perform this simulation 10,000 times on each object. 
Though penalties have been applied to encourage realism, undesired poses still exist. Thus, we design a validation process to select high-quality poses. 
A qualified pose must meet three criteria: (i) the hand must make contact with the object (it may fail to grasp small/thin objects); (ii) no obvious hand-object intersection (the volume of the intersection part should be smaller than a pre-defined threshold); and (iii) no self-penetration for the hand. Finally, to avoid introducing domain gaps, we align the orientation of the generated pose to a random one from the training set and apply perturbation as augmentation.
\RestyleAlgo{ruled}
\SetKwComment{Comment}{/* }{ */}
\begin{algorithm}[t]
\small
\caption{Farthest Pose Sampling Algorithm}\label{alg:fps}
\KwData{Set of poses $\mathbf{P}$, number of sampled poses $M$}
\KwResult{Sampled set of poses $\mathbf{Q}$}
$\mathbf{Q} \gets \varnothing \cup \{\mathbf{v}_0 \in_{\scriptstyle{R}} \mathbf{P}\}$\;
\While{$|\mathbf{Q}| < M$}{
    $\mathbf{v}_{max} \gets \text{None}$\;
    $d_{max} \gets -\infty$\;
    \For{$\mathbf{v}_i \in \mathbf{P},\;\mathbf{v}_i \notin \mathbf{Q}$}{
        $d_i \gets \min_{\mathbf{v}_j \in \mathbf{Q}} D_c(\mathbf{v}_i, \mathbf{v}_j)$\; 
        \If {$d_i > d_{max}$}{
            $d_{max} \gets d_i$\;
            $\mathbf{v}_{max} \gets \mathbf{v}_i$\;
        }
    }
    $\mathbf{Q} \gets \mathbf{Q} \cup \{\mathbf{v}_{max}\}$\;
}
\end{algorithm}

\vspace{-3mm}
\paragraph{Intra-distribution Sampling.}
The grasping pose partly depends on the initial object pose. 
Yet, the number of feasible grasping poses is inherently limited by the shape of the object,~\eg, a bowl can only be placed upright or upside down on a plane, each poses having very limited ways of grasping.
Though it is possible to grasp the same object multiple times, the resulting poses could be too similar to one another, as different grasping poses do not occur with equal probability. 
To promote pose diversity and avoid cases that dominate the distribution, inspired by~\cite{eldar1997farthest}, we propose a farthest-pose-sampling (FPS) strategy to select poses as evenly as possible for each object category.

\begin{figure}[t]
    \centering
    \includegraphics[width=\linewidth]{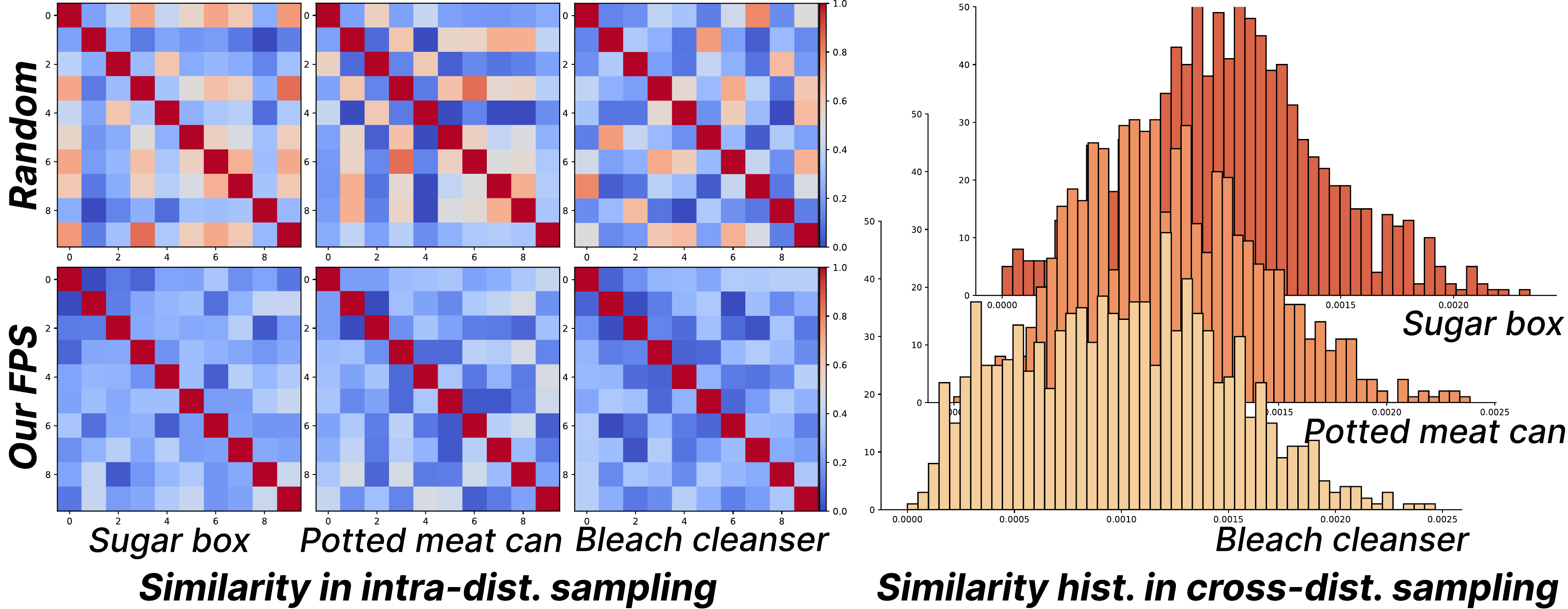}
    \vspace*{-6mm}
    \caption{Visualizations in sampling. For three object categories in HO3D, we show the similarity of sampled real-world poses with our FPS or random sampling (left) and the histograms of similarity between sampled real-world and synthetic poses (right).}
    \label{fig:sampling}
    \vspace{-4mm}
\end{figure}

For real-world and synthetic grasping pose sets $\mathbf{P}_{r}$ and $\mathbf{P}_{s}$, we perform this intra-distribution sampling separately to obtain the sampled sets $\mathbf{Q}_{r}$ and $\mathbf{Q}_{s}$. The subscript is ignored for brevity. The pose vector $\mathbf{v}$ is constructed as
\begin{equation}
\small
    \mathbf{v}=\operatorname{Concat}(\mathbf{h}_s,\mathbf{o}_q,\mathbf{o}_t),\quad \mathbf{v} \in \mathbb{R}^{n}
\end{equation}
where $\mathbf{h}_s$ and $\mathbf{o}_q$ denote the quaternion representation of the hand pose and object rotation, respectively. Note that the global orientation is removed from the grasping pose, which means we compute the distance at the canonical pose.
As described in Alg.~\ref{alg:fps}, FPS begins by initializing the sampled pose set $\mathbf{Q}$ using a random sample $\mathbf{v}_0$ from the input pose set $\mathbf{P}$.
While the number of samples in $\mathbf{Q}$ is less than $M$ ($M$ and $N$ for real-world and synthetic data, respectively), it traverses the remaining poses in $\mathbf{P}$ and selects $\mathbf{v}_{max}$ with the largest nearest distance $d_{max}$ to $\mathbf{Q}$ iteratively. The distance between two pose vectors $\mathbf{v}_i \!\in\! \mathbf{P},\mathbf{v}_i \!\notin\! \mathbf{Q}$ and $\mathbf{v}_j \!\in\! \mathbf{Q}$ is computed using the cosine distance $D_c(\cdot,\cdot)$, \ie,
\begin{equation}
\small
    D_c(\mathbf{v}_i,\mathbf{v}_j)\!=\!\frac{\mathbf{v}_i\cdot \mathbf{v}_j}{||\mathbf{v}_i||||\mathbf{v}_j||}\!=\!\frac{\sum_{k=1}^{n}v_i^kv_j^k}{\sqrt{\sum_{k=1}^{n}{v_i^k}^2}\!\cdot\! \sqrt{\sum_{k=1}^{n}{v_j^k}^2}}.
    \vspace{-1mm}
\end{equation}
As Fig.~\ref{fig:sampling} (left) shows, the sampled poses are more distinctive from each other when applying our FPS strategy.

\begin{table*}[t]
    \centering
    \resizebox{\linewidth}{!}{
        \begin{tabular}{r|l|cccccc|cccccc}
        \toprule
        &\multicolumn{1}{c|}{\multirow{2}{*}{Methods}} & \multicolumn{6}{c|}{\emph{Procrustes Alignment}} & \multicolumn{6}{c}{\emph{Root-relative}} \\
        \cmidrule{3-14}
        & & J-PE $\downarrow$ & J-AUC $\uparrow$   & V-PE $\downarrow$ & V-AUC $\uparrow$ & F@5 $\uparrow$ & F@15 $\uparrow$ & J-PE $\downarrow$ & J-AUC $\uparrow$ & V-PE $\downarrow$ & V-AUC $\uparrow$ & F@5 $\uparrow$ & F@15 $\uparrow$ \\
        \midrule
        &S2Hand~\cite{chen2021model}  & 7.3 & 85.5 & - & - & - & - & - & - & - & - & - & - \\
        &VIBE~\cite{kocabas2020vibe} & 6.4 & 87.1 & - & - & - & - & - & - & - & - & - & - \\
        &TCMR~\cite{choi2021beyond} & 6.3 & 87.5 & - & - & - & - & - & - & - & - & - & - \\
        &H2ONet~\cite{xu2023h2onet} & {5.3} & 89.4 & 5.2 & 89.6 & 80.5 & {99.3} & 13.7 & 74.8 & 12.7 & 76.6 & 52.1 & 92.3 \\
        \multirow{-5}{*}{\rotatebox{90}{Multi-frame}} &Deformer~\cite{fu2023deformer} & {5.2} & {89.6} & - & - & - & - & - & - & - & - & - & - \\
        \midrule
        &METRO~\cite{lin2021end} & 7.0 & - & - & - & - & - & 15.2 & - & - & - & - & - \\
        &Spurr~\etal~\cite{spurr2020weakly} & 6.8 & 86.4 & - & - & - & - & 17.3 & 69.8 & - & - & - & - \\
        &MeshGraphormer~\cite{lin2021mesh} & 6.4 & 87.2 & - & - & - & - & 15.2 & - & - & - & -    & - \\
        &Liu~\etal~\cite{liu2021semi} & 6.6 & - & - & - & - & - & 15.3 & - & - & - & - & - \\
        &Tse~\etal~\cite{tse2022collaborative} & - & - & - & - & - & - & 16.1 & 72.2 & - & - & - & - \\
        \cmidrule{2-14}
        &HandOccNet~\cite{park2022handoccnet} & 5.8 & 88.4 & 5.5 & 89.0 & 78.0 & 99.0 & 14.0 & 74.8 & 13.1 & 76.6 & 51.5 & 92.4 \\
        \rowcolor{highlight}
        \cellcolor{white}&+ Our \ourname{} & \textbf{5.2} & \textbf{89.6} & \textbf{5.0} & \textbf{89.9} & \textbf{81.3} & \textbf{99.2} & \textbf{11.9} & \textbf{77.8} & \textbf{11.5} & \textbf{78.5} & \textbf{55.6} & \textbf{93.6} \\
        \cmidrule{2-14}
        &MobRecon~\cite{chen2022mobrecon}  & 6.4 & 87.3 & 5.6 & 88.9 & 78.5 & 98.8 & 14.2 & 73.7 & 13.1 & 76.1 & 50.8 & 92.1 \\
        \rowcolor{highlight}
        \cellcolor{white}&+ Our \ourname{}  & \textbf{6.2} & \textbf{87.6} & \textbf{5.4 }& \textbf{89.3} & \textbf{79.2} & \textbf{99.1} & \textbf{13.2} & \textbf{74.9} & \textbf{12.3} & \textbf{76.6} & \textbf{51.8} & \textbf{92.5} \\
        \cmidrule{2-14}
        &H2ONet~\cite{xu2023h2onet} & 5.7 & 88.9 & 5.5 & 89.1 & 80.1 & 99.0 & 14.0 & 74.6 & 13.0 & 76.2 & 51.3 & 92.1 \\
        \rowcolor{highlight}
        \cellcolor{white}\multirow{-10}{*}{\rotatebox{90}{Monocular}} &+ Our \ourname{} & \textbf{5.1} & \textbf{89.8} & \textbf{5.1} & \textbf{89.8} & \textbf{81.3} & \textbf{99.2} & \textbf{12.9} & \textbf{76.0} & \textbf{12.5} & \textbf{76.5} & \textbf{52.1} & \textbf{92.2} \\
        \bottomrule
        \end{tabular}}\vspace*{-3mm}
    \caption{Results on ``S0'' (default) data split of DexYCB. -: unavailable results. Our method {\bf \em consistently\/} boosts all three baselines.}
    \label{tab:dexycb_s0}
    \vspace*{-2mm}
\end{table*}

\begin{table*}[t]
    \centering
    \resizebox{\linewidth}{!}{
        \begin{tabular}{l|r@{\text{$\;\;$}}lr@{\text{$\;\;$}}lr@{\text{$\;\;$}}lr@{\text{$\;\;$}}lr@{\text{$\;\;$}}lr@{\text{$\;\;$}}l|r@{\text{$\;\;$}}lr@{\text{$\;\;$}}lr@{\text{$\;\;$}}lr@{\text{$\;\;$}}lr@{\text{$\;\;$}}lr@{\text{$\;\;$}}l}
        \toprule
        \multicolumn{1}{c|}{\multirow{2}{*}{Methods}} & \multicolumn{12}{c|}{\emph{Procrustes Alignment}} & \multicolumn{12}{c}{\emph{Root-relative}} \\
        \cmidrule{2-25}
        & \multicolumn{2}{c}{J-PE $\downarrow$} & \multicolumn{2}{c}{J-AUC $\uparrow$}   & \multicolumn{2}{c}{V-PE $\downarrow$} & \multicolumn{2}{c}{V-AUC $\uparrow$} & \multicolumn{2}{c}{F@5 $\uparrow$} & \multicolumn{2}{c|}{F@15 $\uparrow$} & \multicolumn{2}{c}{J-PE $\downarrow$} & \multicolumn{2}{c}{J-AUC $\uparrow$} & \multicolumn{2}{c}{V-PE $\downarrow$} & \multicolumn{2}{c}{V-AUC $\uparrow$} & \multicolumn{2}{c}{F@5 $\uparrow$} & \multicolumn{2}{c}{F@15 $\uparrow$} \\
        \midrule
        \cite{park2022handoccnet} & 6.6 & & 86.8 & & 6.3 & & 87.3 & & 72.1 & & 98.4 & & 17.4 & & 67.6 & & 16.8 & & 68.5 & & 40.7 & & 86.2 & \\
        \rowcolor{highlight}
        + Ours & 6.1 & \textbf{+0.5} & 87.8 & \textbf{+1.0} & 5.9 & \textbf{+0.4} & 88.3 & \textbf{+1.0} & 75.4 & \textbf{+3.3} & 98.7 & \textbf{+0.3} & 15.8 & \textbf{+1.6} & 70.4 & \textbf{+2.8} & 15.3 & \textbf{+1.5} & 71.3 & \textbf{+2.8} & 44.1 & \textbf{+3.4} & 88.5 & \textbf{+2.3} \\
        \midrule
        \cite{chen2022mobrecon} & 6.9 & & 86.2 & & 6.4 & & 87.2 & & 72.1 & & 98.4 & & 18.3 & & 66.0 & & 17.4 & & 67.5 & & 39.7 & & 85.6 & \\
        \rowcolor{highlight}
        + Ours & 6.6 & \textbf{+0.3} & 86.7 & \textbf{+0.5} & 6.1 & \textbf{+0.3} & 87.9 & \textbf{+0.7} & 74.4 & \textbf{+2.3} & 98.6 & \textbf{+0.2} & 17.5 & \textbf{+0.8} & 67.7 & \textbf{+1.7} & 16.7 & \textbf{+0.7} & 69.1 & \textbf{+1.6} & 41.1 & \textbf{+1.4} & 86.8 & \textbf{+1.2} \\
        \midrule
        \cite{xu2023h2onet} & 6.2 & & 87.7 & & 6.4 & & 87.2 & & 72.4 & & 98.6 & & 17.6 & & 67.9 & & 17.2 & & 68.4 & & 39.4 & & 86.2 & \\
        \rowcolor{highlight}
        + Ours & 6.0 & \textbf{+0.2} & 88.1 & \textbf{+0.4} & 6.0 & \textbf{+0.4} & 88.1 & \textbf{+0.9} & 75.4 & \textbf{+3.0} & 98.7 & \textbf{+0.1} & 17.0 & \textbf{+0.6} & 68.9 & \textbf{+1.0} & 16.4 & \textbf{+0.8} & 69.7 & \textbf{+1.3} & 41.5 & \textbf{+2.1} & 87.2 & \textbf{+1.0} \\
        \bottomrule
        \end{tabular}}\vspace*{-3mm}
    \caption{Results on ``S1'' (unseen subjects) data split of DexYCB. Our method 
    {\bf \em consistently\/} improves all baselines for all metrics.}
    \label{tab:dexycb_s1}
    \vspace*{-4mm}
\end{table*}

\vspace{-3mm}
\paragraph{Cross-distribution Sampling.}
Likewise, the synthetic grasping poses should also be novel and diverse. 
Simply blending them with real-world poses leads to uneven distribution, as shown in Fig.~\ref{fig:sampling} (right), due to the presence of similar samples. 
While our FPS strategy can mitigate this issue, it can not ensure sufficient inclusion of real-world poses, since $M\!\ll\! N$, \eg, $M\!=\!10$ and $N\!=\!500$ for HO3D. 
Additionally, the number of possible unique poses is unknown, making it challenging to define precise criteria for identifying different pose categories.
Thus, we leverage the similarity relationship between the real-world and synthetic poses to adaptively control the sampling probability of the synthetic data. Specifically, for each object category, we calculate the similarity matrix and convert it into a discrete probability distribution $P$ for the sampling process:
\begin{equation}
\small
    \vspace{-1mm}
    P(X\!=\!\mathbf{v}_i^s)=\operatorname{Norm}(\sum_{j=1}^{M}(1\!-\!D_c(\mathbf{v}_i^s,\mathbf{v}_j^r))),
\end{equation}
where $\mathbf{v}_i^s\!\in\! \mathbf{Q}_{s}$ and $\mathbf{v}_j^r\!\in\! \mathbf{Q}_{r}$. $\operatorname{Norm}(\cdot)$ denotes the min-max normalization. The sampling probability of a synthetic pose is inversely related to its similarity to all real-world poses, yielding a more balanced distribution.

\subsection{Hand Mesh Reconstruction}
\label{sec3.4}
In the task of 3D hand-mesh reconstruction from a single RGB image, the goal is to estimate the 3D hand pose and shape. 
To demonstrate the effectiveness of our approach, we adopt it to train two commonly-used pipelines,~\ie, MANO-based and vertex-based methods. 
The former represents the hand using a predefined template and utilizes MANO coefficients to control its pose and shape, which are regressed directly from the extracted features of the input image.
The latter typically regresses the 3D coordinates of hand vertices in a coarse-to-fine manner.
In this work,
we select three recent methods as our baselines: HandOccNet (MANO-based), MobRecon (vertex-based), and H2ONet (vertex-based).
They not only employ different hand-mesh representations but also utilize unique network architectures and scalable input resolutions. 
By applying our synthesized data to these baselines, we can effectively evaluate the generalizability and versatility of our approach.

\section{Experiments}
\label{sec4:exp}

\subsection{Experimental Settings}
\paragraph{Datasets.}
We employ the following commonly-used hand-object benchmark datasets in our experiments.
(i) \textbf{DexYCB}~\cite{chao2021dexycb} (a real-world dataset): we use the default ``S0'' split (train/test: 406,888/78,768 samples) and the more challenging ``S1'' split with unseen subjects (train/test: 7/2 subjects).
(ii) \textbf{HO3D} (version 2)~\cite{hampali2020honnotate} (a real-world dataset): note that evaluation can only be done by submitting results to the official server.
(iii) \textbf{ObMan}~\cite{hasson2019learning} (a rendering-based synthetic dataset): we select it as the competitor of our generated data.
Note that other rendering/generative-based candidates are not available, including YCB-Afford~\cite{corona2020ganhand} (download link not accessible), ArtiBoost~\cite{yang2022artiboost} (data not open-sourced), and HOGAN~\cite{hu2022hand} (missing essential files for training).
To show the improvement in generalizability, we additionally use  \textbf{MOW}~\cite{cao2021reconstructing} as the in-the-wild test set, which provides 512 annotated samples.

\vspace{-4mm}
\paragraph{Evaluation Metrics.}
We adopt common metrics following~\cite{chao2021dexycb, hampali2020honnotate, chen2022mobrecon, park2022handoccnet, xu2023h2onet}.
J-PE/V-PE denotes the root-relative joint/vertex position error, measuring the average Euclidean distance in $mm$ between the predicted and ground-truth 3D hand joint/vertex coordinates. 
J-AUC/V-AUC computes the area under the percentage of correct keypoints (PCK) curve in several error thresholds for joint/vertex.
F@5 and F@15 measure the harmonic mean of the recall and precision for vertex with $5mm$ and $15mm$ thresholds. 
Procrustes Alignment (PA) aligns the orientation, translation, and scale of the estimated results to match the ground truths.
Fréchet Inception Distance~\cite{heusel2017gans} (FID) measures the image fidelity.

\vspace{-4mm}
\paragraph{Implementation Details.}
Adam optimizer~\cite{kingma2014adam} is applied to train the diffusion model and the reconstruction methods.
2D conditions are 256$\times$256 and rendered using PyRender. The diffusion model is trained on 16 NVIDIA 2080Ti GPUs with a batch size of 256 and a learning rate of 0.00008 for 700,000/200,000 iterations on DexYCB/HO3D. Please refer to our supp. material for other details.

\begin{table}[t]
    \centering
    \resizebox{\linewidth}{!}{
        \begin{tabular}{r|l|cccccc}
        \toprule
        & \multicolumn{1}{c|}{Methods} & J-PE $\downarrow$ & J-AUC $\uparrow$   & V-PE $\downarrow$ & V-AUC $\uparrow$  & F@5 $\uparrow$ & F@15 $\uparrow$ \\
        \midrule
        &Hasson~\etal~\cite{hasson2020leveraging} & 11.4 & 77.3 & 11.4 & 77.3 & 42.8 & 93.2 \\
        &Hasson~\etal~\cite{hasson2021towards} & - & - & 14.7 & - & 39.0 & 88.0 \\
        &S2Hand~\cite{chen2021model} & 11.4 & 77.3 & 11.2 & 77.7 & 45.0 & 93.0 \\
        &Liu~\etal~\cite{liu2021semi} & 9.8 & - & 9.4 & 81.2 & 53.0 & 95.7 \\
        & VIBE~\cite{kocabas2020vibe} & 9.9 & - & 9.5 & - & 52.6 & 95.5 \\
        & TCMR~\cite{choi2021beyond} & 11.4 & -  & 10.9 &- & 46.3 & 93.3 \\
        & TempCLR~\cite{ziani2022tempclr} & 10.6 & -  & 10.6 &- & 48.1 & 93.7 \\
        & Deformer~\cite{fu2023deformer} & 9.4 & - & 9.1 & - & 54.6 & 96.3 \\ 
        \multirow{-9}{*}{\rotatebox{90}{Multi-frame}} & H2ONet~\cite{xu2023h2onet} & 8.5 & 82.9 & 8.6 & 82.8 & 57.0 & 96.6 \\
        \midrule
        &Pose2Mesh~\cite{choi2020pose2mesh} & 12.5 & - & 12.7 & - & 44.1 & 90.9 \\
        &I2L-MeshNet~\cite{moon2020i2l} & 11.2 & - & 13.9 & - & 40.9 & 93.2 \\
        &ObMan~\cite{hasson2019learning} & 11.1 & - & 11.0 & 77.8 & 46.0 & 93.0 \\
        &HO3D~\cite{hampali2020honnotate} & 10.7 & 78.8 & 10.6 & 79.0 & 50.6 & 94.2 \\
        &METRO~\cite{lin2021end} & 10.4 & - & 11.1 & - & 48.4 & 94.6 \\
        &Liu~\etal~\cite{liu2021semi} & 10.2  & 79.7 & 9.8 & 80.4 & 52.9 & 95.0 \\
        &I2UV-HandNet~\cite{chen2021i2uv} & 9.9 & 80.4 & 10.1 & 79.9 & 50.0 & 94.3 \\   
        &Tse~\etal~\cite{tse2022collaborative} & - & - & 10.9 & - & 48.5 & 94.3 \\
        &AMVUR~\cite{jiang2023probabilistic} & {8.3} & {83.5} & {8.2} & {83.6} & {60.8} & 96.5 \\
        \cmidrule{2-8}
        &HandOccNet~\cite{park2022handoccnet} & 9.1 & 81.9 & 9.0 & 81.9 & 56.1 & 96.2 \\
        &HandOccNet*~\cite{park2022handoccnet} & 9.6 & 80.8 & 9.6 & 80.7 & 52.4 & 95.4 \\
        \rowcolor{highlight}
        \cellcolor{white}&+ Our \ourname{} & \textbf{8.5} & \textbf{82.9} & \textbf{8.6} & \textbf{82.9} & \textbf{57.7} & \textbf{97.2} \\
        \cmidrule{2-8}
        &MobRecon~\cite{chen2022mobrecon}           & 9.4   & 81.3    & 9.5   & 81.0    & 53.3 & 95.5 \\
        &MobRecon$^\dagger$~\cite{chen2022mobrecon} & 9.2 & 81.6 & 9.4 & 81.2 & 53.8 & 95.7 \\
        \rowcolor{highlight}
        \cellcolor{white}&+ Our \ourname{} & \textbf{8.7} & \textbf{82.6} & \textbf{8.8} & \textbf{82.5} & \textbf{56.1} & \textbf{97.0} \\
        \cmidrule{2-8}
        & H2ONet~\cite{xu2023h2onet} & 9.0 & 82.0 & 9.0 & 81.9 & 55.4 & 96.0  \\
        \rowcolor{highlight}
        \cellcolor{white}\multirow{-16}{*}{\rotatebox{90}{Monocular}} & + Our \ourname{} & \textbf{8.2} & \textbf{83.6} & \textbf{8.4} & \textbf{83.2} & \textbf{58.5} & \textbf{97.2} \\
        \bottomrule
        \end{tabular}
    }
    \vspace*{-3mm}
    \caption{Results on HO3D (\emph{Procrustes Alignment}). $^*$: reproduced results using its official code. $^\dagger$: complement data is used. -: unavailable results.
    Our method brings improvement consistently.}
    \label{tab:ho3d_after_pa}
    \vspace{-4mm}
\end{table}

\subsection{Comparison with State-of-the-art Methods}
\begin{figure}[t]
    \centering
    \includegraphics[width=\linewidth]{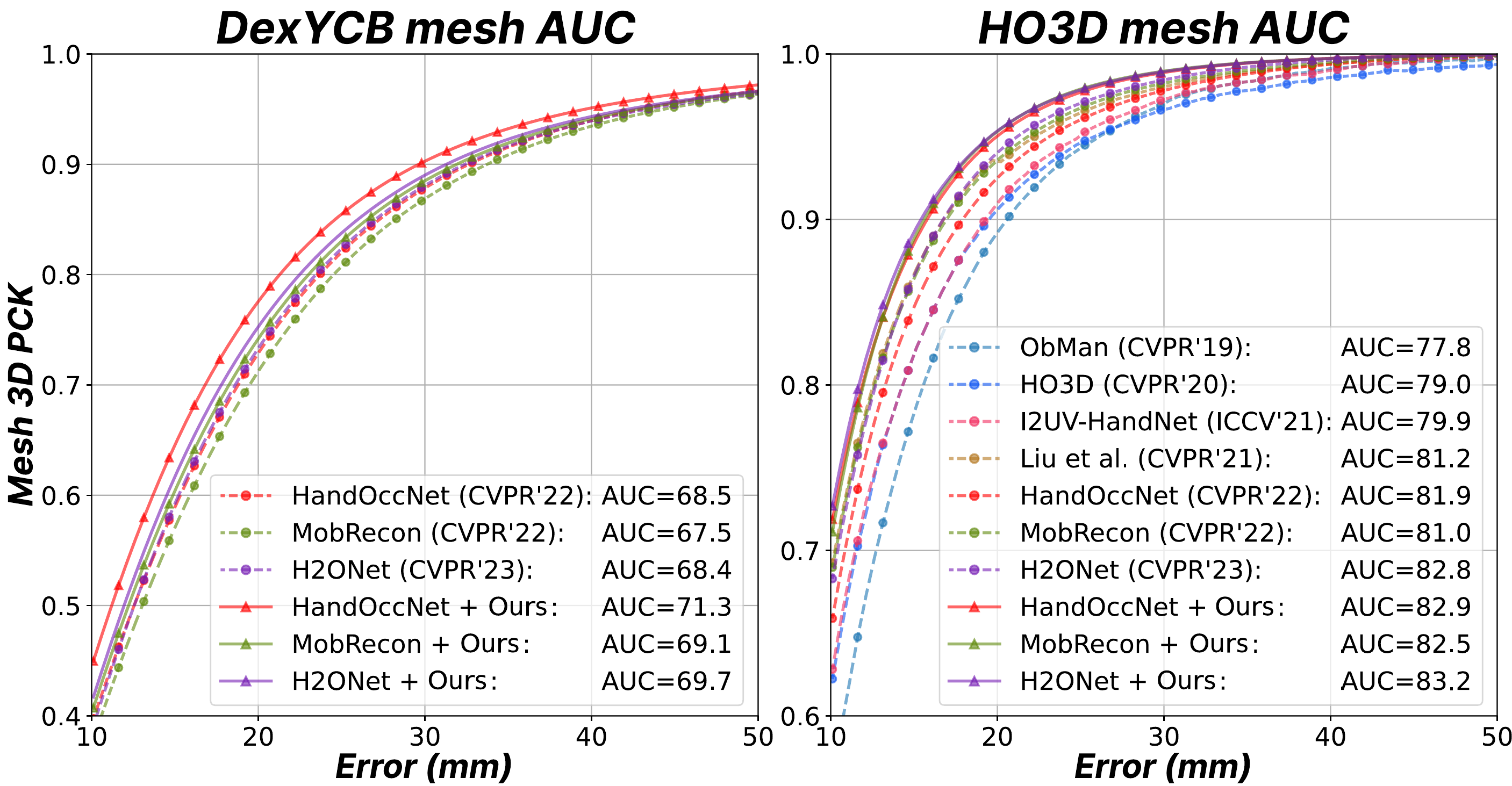}
    \vspace*{-7mm}
    \caption{The mesh AUC comparison under different thresholds. All baselines are improved consistently after applying our data.}
    \label{fig:pck}
    \vspace{-4mm}
\end{figure}

\paragraph{Evaluation on DexYCB.}
We first conduct quantitative comparisons on DexYCB. 
To demonstrate the effectiveness of our method, we evaluate metrics before and after performing PA,
as presented in Tab.~\ref{tab:dexycb_s0} and Tab.~\ref{tab:dexycb_s1} for ``S0'' and ``S1'' data splits, respectively. 
Additionally, Fig.~\ref{fig:pck} (left) visualizes the root-relative mesh PCK/AUC comparison for the more challenging ``S1'' split.
Our synthetic data significantly improves the performance of HandOccNet~\cite{park2022handoccnet} and the monocular-based H2ONet~\cite{xu2023h2onet}, enabling them even surpass some multi-frame methods, such as Deformer~\cite{fu2023deformer}.
MobRecon~\cite{chen2022mobrecon} exhibits relatively small performance gains due to its mobile-friendly designs. Its capability is limited by its lower FLOPs (0.46G) and the number of parameters (8.23M) compared to H2ONet (0.74G/25.88M) and HandOccNet (15.47G/37.22M).
Our \ourname{} consistently boosts the performance of all three baselines across all metrics and two data splits, giving strong evidence of its effectiveness.
It even outperforms multi-frame methods (see Tab.~\ref{tab:dexycb_s0}), by taking only a single view as its input.

\begin{figure*}[t]
    \centering
    \includegraphics[width=0.97\linewidth]{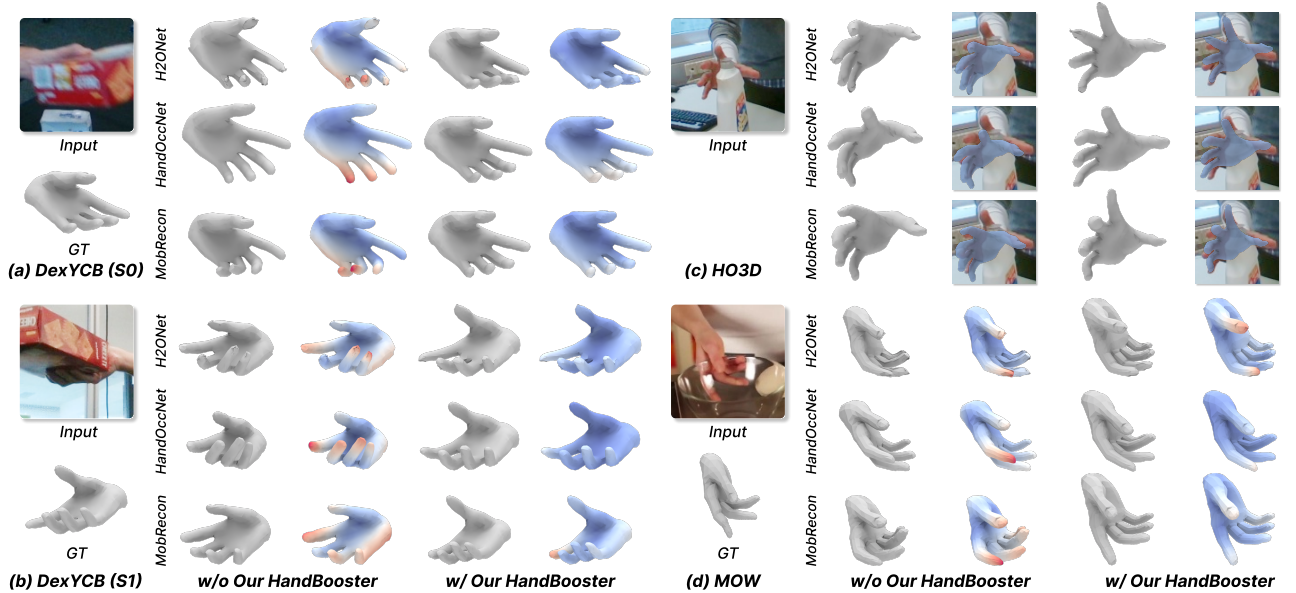}
    \vspace*{-3mm}
    \caption{Qualitative comparison of~\cite{xu2023h2onet, park2022handoccnet, chen2022mobrecon} w/ and w/o our \ourname{}. (For each case, left: predicted results, right: error map.)}
    \label{fig:qualitative_results}
    \vspace{-4mm}
\end{figure*}

\begin{figure}[t]
    \centering
    \includegraphics[width=0.96\linewidth]{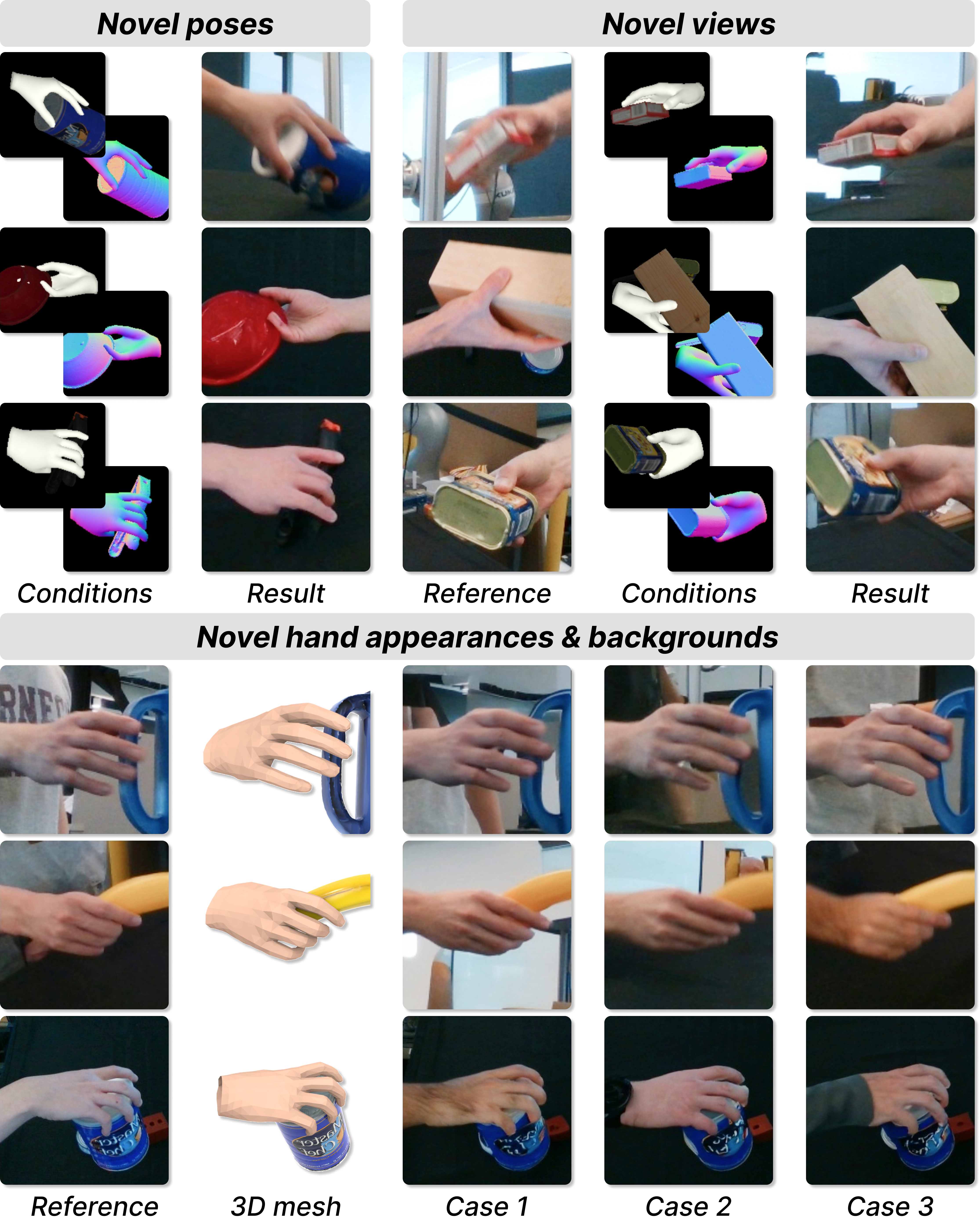}
    \vspace{-3mm}
    \caption{Generated examples for novel views/poses with realistic hand appearance and backgrounds.}
    \label{fig:dm_output}
    \vspace{-5mm}
\end{figure}

Fig.~\ref{fig:qualitative_results} (a-b) and (d) show qualitative results on DexYCB. Note that we directly use models pre-trained on DexYCB's ``S0'' split to test on MOW. 
Comparing the results w/ and w/o using our data, the performance gains clearly show \ourname{}'s efficacy and generalizability. Fig.~\ref{fig:dm_output} shows some of the generated samples with novel views/poses, where the fine-grained alignment with conditions and the realistic appearance show the controllability of \ourname{}. More results are shown in the supp. material.

\vspace{-4mm}
\paragraph{Evaluation on HO3D.}
We perform the 
same experiments on HO3D. 
Since the ground truths of the test set are not publicly available, some results are from previous papers and the official evaluation server.
Tabs.~\ref{tab:ho3d_after_pa} and~\ref{tab:ho3d_before_pa} show results before and after PA, respectively.
To facilitate comparison, we also provide the mesh PCK/AUC after PA, visualized in Fig.~\ref{fig:pck} (right). It is clear that all our baselines consistently achieve improved precision on all the metrics when boosted by our generated data, showing the great effectiveness and robustness of our method.
For an intuitive comparison, we present visual comparisons before and after applying our method in Fig.~\ref{fig:qualitative_results} (c).
Benefiting from the diverse and novel samples in our generated data, all these baselines become able to produce plausible shapes and estimate accurate hand orientations, even under severe occlusions in the inputs.

\begin{table}[t]
    \centering
    \resizebox{\linewidth}{!}{
        \begin{tabular}{l|r@{\text{$\;\;$}}lr@{\text{$\;\;$}}lr@{\text{$\;\;$}}lr@{\text{$\;\;$}}lr@{\text{$\;\;$}}lr@{\text{$\;\;$}}l}
        \toprule
        \multicolumn{1}{c|}{Methods} & \multicolumn{2}{c}{J-PE $\downarrow$} & \multicolumn{2}{c}{J-AUC $\uparrow$}   & \multicolumn{2}{c}{V-PE $\downarrow$} & \multicolumn{2}{c}{V-AUC $\uparrow$} & \multicolumn{2}{c}{F@5 $\uparrow$} & \multicolumn{2}{c}{F@15 $\uparrow$} \\
        \midrule
        \cite{park2022handoccnet} & 24.9 & & 53.9 & & 24.2 & & 55.1 & & 26.0 & & 72.9 & \\
        \cite{park2022handoccnet}* & 25.1 & & 53.3 & & 24.5 & & 54.4 & & 25.6 & & 72.8 & \\
        \rowcolor{highlight}
        + Ours & 21.1 & \textbf{+4.0} & 59.4 & \textbf{+6.1} & 20.5 & \textbf{+4.0} & 60.4 & \textbf{+6.0} & 28.7 & \textbf{+3.1} & 77.9 & \textbf{+5.1} \\
        \midrule
        \cite{chen2022mobrecon} & 25.2 & & 53.7 & & 24.4 & & 55.0 & & 26.4 & & 72.0 & \\
        \rowcolor{highlight}
        + Ours & {23.4} & \textbf{+1.8} & 56.5 & \textbf{+2.8} & {22.6} & \textbf{+1.8} & {57.9} & \textbf{+2.9} & {27.7} & \textbf{+1.3} & {75.3} & \textbf{+3.3} \\
        \midrule
        \cite{xu2023h2onet} & 26.3 & & 52.3 & & 25.5 & & 53.5 & & 24.9 & & 71.5 & \\
        \rowcolor{highlight}
        + Ours & 24.0 & \textbf{+2.3} & {56.7} & \textbf{+4.4} & 23.3 & \textbf{+2.2} & 57.7 & \textbf{+4.2} & 26.6 & \textbf{+1.7} & 74.4 & \textbf{+2.9} \\
        \bottomrule
        \end{tabular}}
    \vspace*{-3mm}
    \caption{Results on HO3D (\emph{Root-relative}). $^*$: reproduced results using its official code. Our method boosts baselines significantly.
    }
    \label{tab:ho3d_before_pa}
    \vspace{-1mm}
\end{table}

\subsection{Ablation Studies}
We conduct ablation studies on DexYCB to evaluate the effectiveness of \ourname{}, as shown in Tab.~\ref{tab:abl_study}. MobRecon is selected as the baseline due to its fast training speed.

\vspace{-3.5mm}
\paragraph{Rendering-based Synthesis.}
We show the importance of realism in data generation by comparing our data with ObMan, where hand appearance, grasping pose, and background are randomly selected for rendering. 
Comparing Rows (a) and (b), introducing ObMan only slightly improves the root-relative metrics after PA.
Correspondingly, comparing Rows (a) and (f), our data enhances performance on all metrics even without the adoption of our other techniques, revealing the importance of realism in the data generation. 
Further, we compute the FID score between our data/ObMan and the training set of DexYCB. The scores of 2.91/8.52 reflect the realism of our generated data.

\vspace{-3.5mm}
\paragraph{Content-aware Conditions.}
To better understand the influence of different conditions for 3D hand-mesh reconstruction, we evaluate various candidate combinations. 
Due to the lack of shape information, the skeleton and foreground segmentation are not used in this analysis. 
The results, shown in Rows (d-f) of Tab.~\ref{tab:abl_study}, reveal that incorporating both normal and texture yields the highest performance.
Substituting either component with alternative candidates leads to a significant decrease in performance, thus supporting our idea of constructing conditions to be both informative and easy to understand.
Further, by introducing embedded hand orientation as an additional control, our model shows noticeable improvements in root-relative metrics, further validating the effectiveness of our design.

\vspace{-3mm}
\paragraph{Novel Condition Creator.}
We also investigate the impact of components in our novel condition creator. 
Comparing Rows (f) and (g), though only utilizing novel grasping poses boosts root-relative performance, certain poses may dominate the entire distribution and limit the performance. 
Comparing Rows (h-i) with (g), performing the intra- and cross-distribution sampling sequentially brings boosts across all metrics clearly, demonstrating their necessities.
Further experimental results are presented in the supp. material.

\begin{table}[t]
    \centering
    \resizebox{\linewidth}{!}{%
    \begin{tabular}{c|l|cc|cc}
        \toprule
        &\multicolumn{1}{c|}{\multirow{2}{*}{Models}} & \multicolumn{2}{c|}{\emph{Root-relative}} & \multicolumn{2}{c}{\emph{Procrustes Align.}} \\
        \cmidrule{3-6}
        & & J-PE $\downarrow$ & V-PE $\downarrow$ & J-PE $\downarrow$ & V-PE $\downarrow$ \\
        \midrule
        (a) & Baseline & 14.20 & 13.05 & 6.36 & 5.59  \\
        (b) & w/ ObMan & 14.00 & 13.00 & 6.40 & 5.60 \\
        \midrule
        (c) & w/ Depth \& Segment. & 14.91 & 13.94 & 6.64 & 5.85  \\
        (d) & w/ Normal \& Segment. & 14.82 & 13.85 & 6.60 & 5.81  \\
        (e) & w/ Normal \& Texture & 13.93 & 12.92 & 6.37 & 5.53  \\
        (f) & + Embedded Orientation & 13.79 & 12.78 & 6.33 & 5.49 \\
        \midrule
        (g) & + Novel Conditions & 13.51 & 12.57 & 6.34 & 5.55 \\
        (h) & + Intra-dist. Sampling & 13.42 & 12.42 & 6.28 & 5.46 \\
        (i) & + Cross-dist. Sampling & \best{13.25} & \best{12.34} & \best{6.20} & \best{5.36} \\
         \bottomrule
    \end{tabular}%
    }
    \vspace*{-3mm}
    \caption{Ablation study on major components.}
    \label{tab:abl_study}
    \vspace{-6mm}
\end{table}

\section{Conclusion}
\label{sec5:conclusion}

We presented \ourname{}, a new generative method to boost 3D hand-mesh reconstruction by enhancing the data diversity.
First, we create a conditional generative space, from which we can controllably produce realistic and diverse hand-object images with reliable 3D annotations.
Then, we explore this space to produce novel and diverse training samples by formulating a novel condition creator and two similarity-aware sampling strategies.
Extensive experiments on three baselines and two common benchmarks demonstrate our effectiveness and SOTA performance.

\vspace{-3mm}
\paragraph{Acknowledgments}
This work was supported by the Research Grants Council of the Hong Kong Special Administrative Region, China (Project No. T45-401/22-N and No. CUHK 14201921) and the National Natural Science Foundation of China under grant No. 62372091. Hao Xu thanks for the care \& support from Yutong Zhang \& his family.
\clearpage


\end{document}